\documentclass{article}

\usepackage{microtype}
\usepackage{graphicx}
\usepackage{subfig}
\usepackage{floatrow}
\usepackage{verbatim}
\usepackage{multirow}
\usepackage[export]{adjustbox}
\usepackage{amssymb}
\usepackage{booktabs} 
\usepackage[dvipsnames]{xcolor}
\usepackage{pgfplots}
\pgfplotsset{compat=1.17}
\usepgfplotslibrary{groupplots}
\usepackage{arydshln}
\usepackage{setspace}
\usepackage{tikz}
\usetikzlibrary{positioning,calc, matrix}
\usepackage{dblfloatfix}
\usepackage{hyperref}

\usepackage{amsmath} 
\usepackage{amsfonts} 
\usepackage{bm}

\def\eqref#1{equation~\ref{#1}}

\def\1{\bm{1}}

\def\rva{{\mathbf{a}}}
\def\rvb{{\mathbf{b}}}
\def\rvc{{\mathbf{c}}}

\def\rvi{{\mathbf{i}}}
\def\rvo{{\mathbf{o}}}

\def\rvx{{\mathbf{x}}}
\def\rvy{{\mathbf{y}}}

\def\rvxpi{{\boldsymbol{\pi}}}

\def\mE{{\bm{E}}}

\def\mX{{\bm{X}}}

\DeclareMathAlphabet{\mathsfit}{\encodingdefault}{\sfdefault}{m}{sl}
\SetMathAlphabet{\mathsfit}{bold}{\encodingdefault}{\sfdefault}{bx}{n}

\def\gA{{\mathcal{A}}}
\def\gB{{\mathcal{B}}}

\def\gD{{\mathcal{D}}}
\def\gE{{\mathcal{E}}}

\def\gL{{\mathcal{L}}}

\def\gT{{\mathcal{T}}}

\def\gY{{\mathcal{Y}}}

\def\sR{{\mathbb{R}}}

\def\emL{{L}}

\pgfplotsset{
    bar group size/.style 2 args={
        /pgf/bar shift={%
                -0.5*(#2*\pgfplotbarwidth + (#2-1)*\pgfkeysvalueof{/pgfplots/bar group skip})  + 
                (.5+#1)*\pgfplotbarwidth + #1*\pgfkeysvalueof{/pgfplots/bar group skip}},%
    },
    bar group skip/.initial=2pt,
    plot 0/.style={blue,fill=blue!30!white,mark=none},%
    plot 1/.style={red,fill=red!30!white,mark=none},%
    plot 2/.style={brown!60!black,fill=brown!30!white,mark=none},%
}

\tikzset{
    every left delimiter/.style={xshift=1ex},
    every right delimiter/.style={xshift=-1ex},
    bmatrix/.style={matrix of math nodes,
        inner sep=0pt,
        left delimiter={[},
        right delimiter={]},
        nodes={anchor=center, inner sep=.2333em},
        }
}

\usepackage[accepted]{icml2022}

\icmltitlerunning{STC: Sequence Classification with Partially Labeled Data}

\begin{document}

\twocolumn[
\icmltitle{Star Temporal Classification: \\ Sequence Classification with Partially Labeled Data}



\icmlsetsymbol{equal}{*}

\begin{icmlauthorlist}

\icmlauthor{Vineel Pratap}{fb}
\icmlauthor{Awni Hannun}{zoom}
\icmlauthor{Gabriel Synnaeve}{fb}
\icmlauthor{Ronan Collobert}{apple}
\end{icmlauthorlist}

\icmlaffiliation{fb}{Meta AI}
\icmlaffiliation{zoom}{Zoom AI}
\icmlaffiliation{apple}{Apple, work done while at Meta AI}

\icmlcorrespondingauthor{Vineel Pratap}{vineelkpratap@fb.com}

\icmlkeywords{Machine Learning, ICML}
\vskip 0.3in
]



\printAffiliationsAndNotice{} 

\newcommand{\note}[1]{\textcolor{red}{ #1}}

\begin{abstract}
We develop an algorithm which can learn from partially labeled and unsegmented sequential data. Most sequential loss functions, such as Connectionist Temporal Classification (CTC), break down when many labels are missing. We address this problem with Star Temporal Classification (STC) which uses a special \textit{star} token to allow alignments which include all possible tokens whenever a token could be missing. We express STC as the composition of weighted finite-state transducers (WFSTs) and use GTN (a framework for automatic differentiation with WFSTs) to compute gradients. We perform extensive experiments on automatic speech recognition. These experiments show that STC can recover most of the performance of supervised baseline when up to 70\% of the labels are missing. We also perform experiments in handwriting recognition to show that our method easily applies to other sequence classification tasks.
\end{abstract}

\section{Introduction}
\label{intro}

\begin{figure}[ht]
\centering
\includegraphics[width=\textwidth]{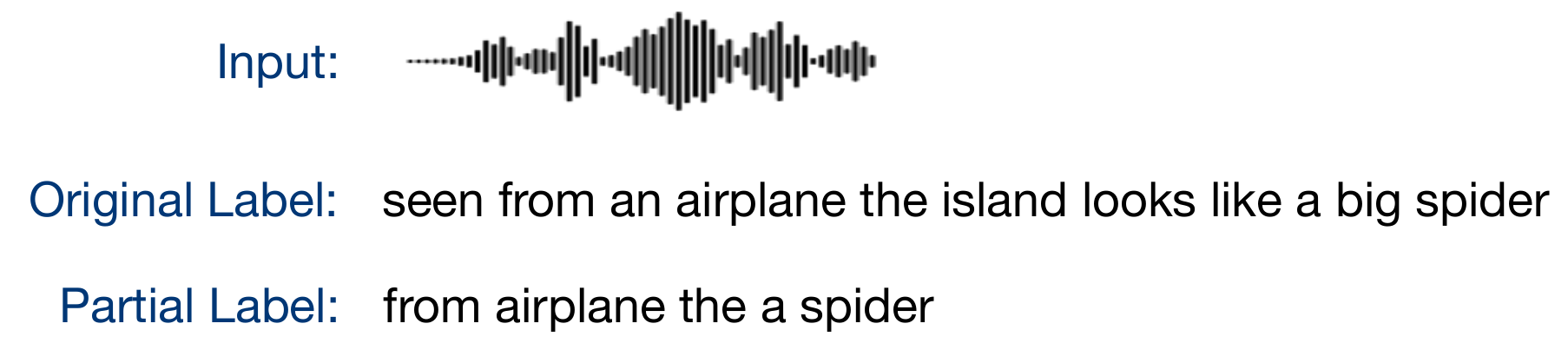}
\caption{An example of speech with a complete and a partial label.}
 \label{fig:pl_example}
 \vspace{-0.1in}
\end{figure}

Applications of machine learning in settings with little or no labeled data are growing rapidly. Much prior work in temporal classification focuses on semi-supervised or self-supervised learning \cite{chapelle2009semi, synnaeve2020endtoend}. Instead, we focus on weakly supervised learning, in which the learner receives incomplete labels for the examples in the data set. Examples include multi-instance learning \cite{Amores2013MultipleIC,zhou2012multi}, and partial-label learning \cite{Jin2002LearningWM, Courpll, pmlr-v32-liug14}. In our case, we assume that each example in the training set is partially labeled, as in Figure~\ref{fig:pl_example}: neither the number of missing words nor their positions in the label sequence are known in advance.

Partial labelling problem could arise in many practical scenarios. In semi-supervised learning or transfer learning of sequence labeling tasks, we can only consider high confidence tokens in the label sequence and discard rest of the tokens. These labels with incomplete tokens can be treated as partial labels. When transcribing a dataset, we can randomly label parts of the dataset and use them as partial labels, at a much lower cost than transcribing the full dataset. We can also consider semi-automatic ways to label a dataset. For example, transcribing a speech dataset could be done by considering only high confidence words from multiple noisy sources available at disposal - keyword spotting models, human raters and ASR models which can be done at a lower overall cost compared to producing high quality transcripts with human raters. We also often face partial alignment problem~\cite{anonymous2022wctc} in practice where beginning and ending tokens are missing. This is a sub-problem of the more general version of partial labelling problem that we solve in this work.   


To solve this weak supervision problem we develop the Star Temporal Classification (STC) loss function. The STC algorithm explicitly accounts for the possibility of missing labels in the output sequence at any position, through a special \textit{star} token. For the example of Figure~\ref{fig:pl_example}, the STC label sequence is \texttt{* from * airplane * the * a * spider *} (where \texttt{*} can be anything but the following label token). We show how to simply implement STC in the weighted finite-state transducer (WFST) framework. Moreover, through the use of automatic differentiation with WFSTs, gradients of the loss function with respect to the inputs are computed automatically, allowing STC to be used with standard neural network training pipelines. 


For practical applications, a naive implementation of STC using WFSTs is intractably slow. We propose several optimizations to improve the efficiency of the STC loss. The \textit{star} token we use results in much smaller graphs since it combines many arcs into one. The \textit{star} token also enables us to substantially reduce the amount memory transfers between the GPU, on which the network is computed, and the CPU, on which the STC algorithm is evaluated. These optimizations result in only a minor increase in model training time when using the STC loss over a CTC baseline.

The STC loss can be applied to most sequence transduction tasks including speech recognition, handwriting recognition, machine translation, and action recognition from videos. Here we demonstrate the effectiveness of STC on both automatic speech recognition and offline handwriting recognition. In speech recognition, with up to 70\% of the labels missing, STC is able to achieve low word error rates. Similarly in handwriting recognition STC can achieve low character error rates with up to 50\% of the labels missing. In both cases STC yields dramatic gains over a baseline system which does not explicitly handle missing labels.

To summarize, the main contributions of this work are:

\begin{itemize}
 \item  We introduce STC, a novel temporal classification algorithm which explicitly handles an arbitrary number of missing labels with an indeterminate location.
 \item We simplify the implementation of STC by using WFSTs and automatic differentiation.
 \item We describe key optimizations for STC enabling it to be used with minimal increase in training time.
 \item We show in practical speech and handwriting recognition tasks that STC yields low word and character error rates with up to 70\% of the labels missing.
\end{itemize}

\section{Related Work}

Many generalizations and variations of CTC have been proposed. Graph-based temporal classification (GTC) \cite{moritz2021semi} generalizes CTC to allow for an $n$-best list of possible labels. \citet{wigington2019multi} also extends CTC for use with for multiple labels. \citet{laptev2021ctc} propose selfless-CTC which disallows self-loops for non-blank output tokens. Unlike these alternatives, STC allows for an arbitrary number of missing tokens anywhere in the label for a given example. A recent work, concurrent with this work, proposes a wild card version of CTC~\cite{anonymous2022wctc}. However, in that work the wild cards, or missing tokens, can only be at the beginning or end of the label.

Extended Connectionist Temporal Classification (ECTC)~\cite{huang2016connectionist} was developed for weakly supervised video action labeling. However, the weak supervision in this case refers to the lack of a segmentation, which is the common paradigm for CTC. The extensions to CTC are intended to handle the much higher ratio of input video frames to output action labels. \citet{dufraux2019lead2gold} develop a sequence level loss function which can learn from noisy labels. They extend the auto segmentation criterion~\cite{collobert2016wav2letter} to include a noise model which explicitly handles up to one missing label between tokens in the output. Unlike that work, STC extends the more commonly used CTC loss and allows for any number of missing tokens.

The idea of expressing CTC as the composition of WFSTs is well known~\cite{miao2015eesen, laptev2021ctc, xiang2019crf}. However, recently developed WFST-based frameworks which support automatic differentiation~\cite{hannun2020differentiable, k2} make the development of variations of the CTC loss much simpler.

\section{Method}
\subsection{Problem Description}
We consider the problem of temporal classification \cite{Kadous2002TemporalCE}, which predicts an output sequence $\rvy = [y_1,....,y_U] \in \gA^{1 \times U}$, where $\gA$ is a fixed alphabet of possible output tokens, from an unsegmented input sequence $\rvx = [x_1,...,x_{T}] \in \sR^{d \times T}$, where each $x_i$ is a $d$-dimensional feature vector.  We also assume that the length of the output $U$ is less than or equal to that of the input, $T$. A partially labeled output sequence $\tilde{\rvy}$ of length $\tilde{U}$, such that ($\tilde{U} \leq U$), is formed by removing zero or more elements from the true output sequence $\rvy$. Given a training set $\gD \triangleq \{ \rvx^{i}, \tilde{\rvy}^{i}  : i= 1, \ldots, N \}$ consisting of input sequences  $\rvx^i = [x_1, \ldots ,x_{T^i}]$ and partially labeled output sequences  $\tilde{\rvy}^i = [y_1, \ldots ,y_{\tilde{U}^i}]$ our goal is to learn temporal models which can predict true output sequences $\rvy^i$. We consider this a weakly supervised temporal classification problem as the learning algorithm only has access to a subset of the true output labels during training. 

\subsection{Weighted Finite-State Transducers}
\begin{figure}[ht]
\centering
\begin{floatrow}
\subfloat[{\scriptsize Example WFST, $\gT_1$\label{fig:gg}}]{
    \raisebox{10pt}{\includegraphics[width=0.46\textwidth]{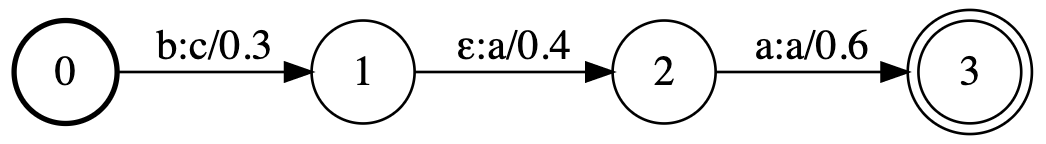}}
}

\subfloat[{\scriptsize Example WFST, $\gT_2$\label{fig:gf}}]{
    \includegraphics[width=0.44\textwidth]{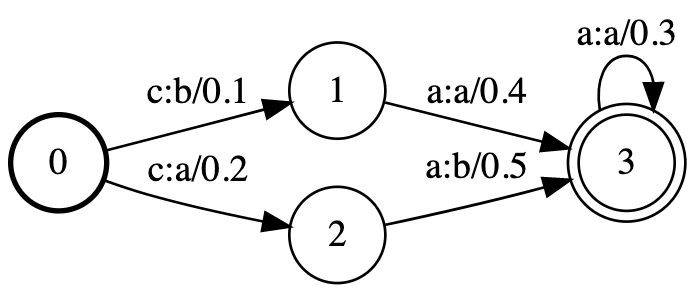}}
\end{floatrow}
\begin{floatrow}
\subfloat[{\scriptsize Composition, $\gT_1 \circ \gT_2 \label{fig:gc}$}]{
    \includegraphics[width=0.55\textwidth]{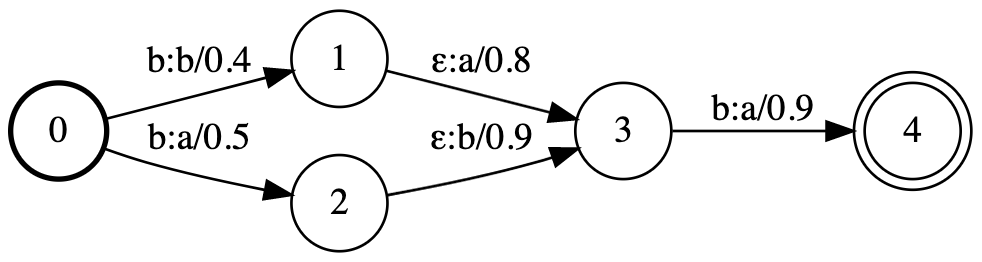}}
\subfloat[{\scriptsize Forward Score, $ \textit{Fwd}(\gT_1) \label{fig:gv}$}]{
\hspace{0.03\textwidth}
\raisebox{7pt}{\includegraphics[width=0.25\textwidth]{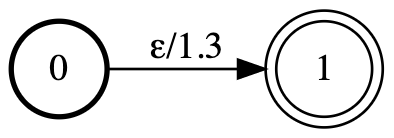}}
\hspace{0.03\textwidth}
}
\end{floatrow}
\caption{Examples of WFSTs and their operations (in log semiring). The arc label “$p$:$r$/$w$” denotes an input label $p$, an output label $r$ and weight $w$. Special symbol, $\epsilon$ allows to make a transition without consuming an input label or without producing an output label. Start states are represented by bold circles and final states by concentric circles.}


 \label{fig:wfst_graphs}
\vskip -0.1in
\end{figure}

A weighted finite-state transducer (WFST) is a generalization of a finite-state automaton (FSA) \cite{mohri2009weighted, mohri2008speech, vidal2005probabilistic} where each transition has an input label from an alphabet $\Sigma$ an
output label from an alphabet $\Delta$ and scalar weight $w$. Figure \ref{fig:gg} shows an example WFST with nodes representing states and arcs representing transitions. A path from an initial to a final state encodes a mapping from an input sequence $\rvi \in \Sigma^*$ to an output sequence $\rvo \in \Delta^*$ and a corresponding score. While WFST operations can be performed in any semiring, in this work we only use the log semiring.

In this work we use composition and shortest distance operations on WFSTs. The composition operation combines WFSTs from different modalities. Given two WFSTs $\gT_1$ and $\gT_2$, if $\gT_1$ transduces $\rva$ to $\rvb$ with weight $w_1$ and  $\gT_2$ transduces  $\rvb$ to  $\rvc$ with weight $w_2$, then their composition $\gT_1 \circ \gT_2$ transduces $\rva$ to $\rvc$  with weight $w_1 + w_2$. The forward score operation is the shortest distance from a start state to a final state in the log semiring. Given a transducer $\gT_1$, the forward score is the log-sum-exp of the scores of all paths from any start state to any final state. The example output graphs from composition and forward score operations are shown in Figure \ref{fig:gc} and Figure \ref{fig:gv} respectively. 

\subsubsection{Autograd with WFSTs}
 \label{sec:autograd}
Most operations on WFST are differentiable with respect to the arc weights of the input graphs. This allows WFSTs to be used dynamically to train neural networks. Frameworks like GTN~\cite{hannun2020differentiable} and k2~\cite{k2} implement automatic differentiation with WFSTs. For the purpose of this work, we use the GTN framework.

\subsection{Connectionist Temporal Classification (CTC)}
Connectionist Temporal Classification (CTC)~\cite{graves2006connectionist} is a widely used loss function for training sequence models where the alignment of input sequence with the target labels is not known in advance.

Consider an input sequence $\rvx = [x_1,\ldots,x_{T}]$ and target sequence $\rvy = [y_1,\ldots,y_U] \in \gA^{1 \times U}$, where $\gA$ is a finite alphabet of possible output tokens and $U \leq T$. The CTC loss uses a special \textit{blank} token, $\langle b \rangle$, to represents frames which do not correspond to an output token. An alignment between the input $\rvx$ and output $\rvy$ in CTC is represented by a sequence of length $T$;  $\rvxpi =[\pi_1, \pi_2, ..., \pi_T]$ where $\pi_t \in \gA \cup \{\textit{blank}\}$. The CTC collapse function $\gB$ maps an alignment to an output, $\gB(\pi) = \rvy$. The function $\gB$ removes all but one of any consecutively repeated tokens and then removes \textit{blank} tokens e.g., (a, b, \textit{blank}, \textit{blank}, b, b, \textit{blank}, a) $\mapsto$ (a, b, b, a). Given the input  $\rvx$, each output is independent of all other outputs, hence the probability of an alignment $P(\rvxpi \vert \rvx)$ is: 
\begin{equation}
    P(\rvxpi \vert \rvx) = \prod_{i=1}^{T} P(\pi_t \vert \rvx).
\end{equation}
There can be many possible alignments $\rvxpi$ for a given $\rvx$ and $\rvy$ pair. The CTC loss computes the negative log probability of $\rvy$ given $\rvx$ by marginalizing over all possible alignments: 
\begin{equation}
    \emL_{CTC} = - \log P(\rvy \vert \rvx) = - \log \sum_{\rvxpi \in \gB^{-1}(\rvy)} P(\rvxpi \vert \rvx).
\label{eq:ctc_loss}
\end{equation}
The sum over all alignments be computed efficiently with dynamic programming as described by \citet{graves2006connectionist}.

\begin{figure*}
\begin{center}
\begin{floatrow}
\subfloat[Emission Graph $\gE_{\mX}$\label{fig:ctc_emissions}]{
    \includegraphics[width=0.45\linewidth]{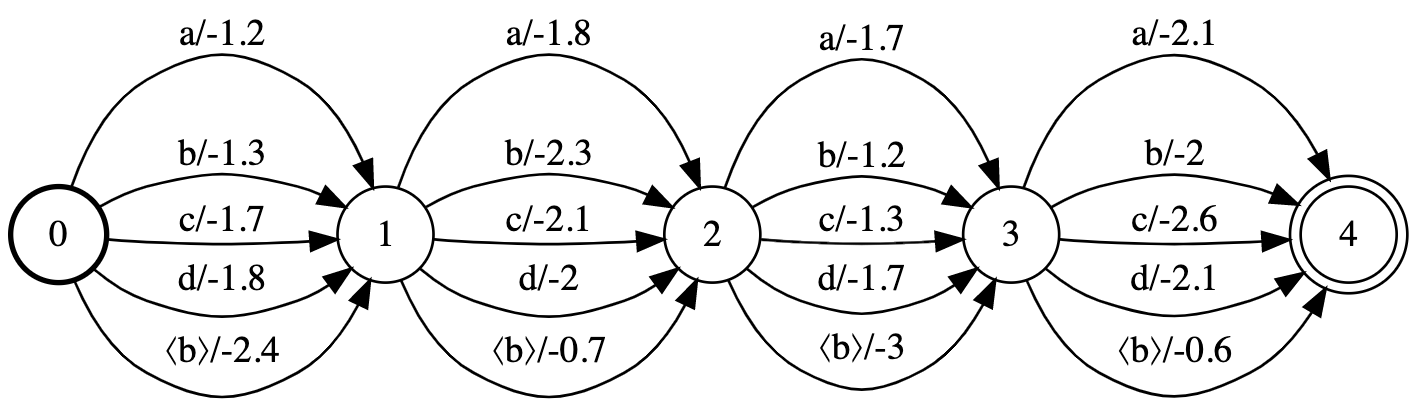}}

\subfloat[Label Graph $\gY_{ctc}$\label{fig:ctc_alignments}]{
    \includegraphics[width=0.5\linewidth]{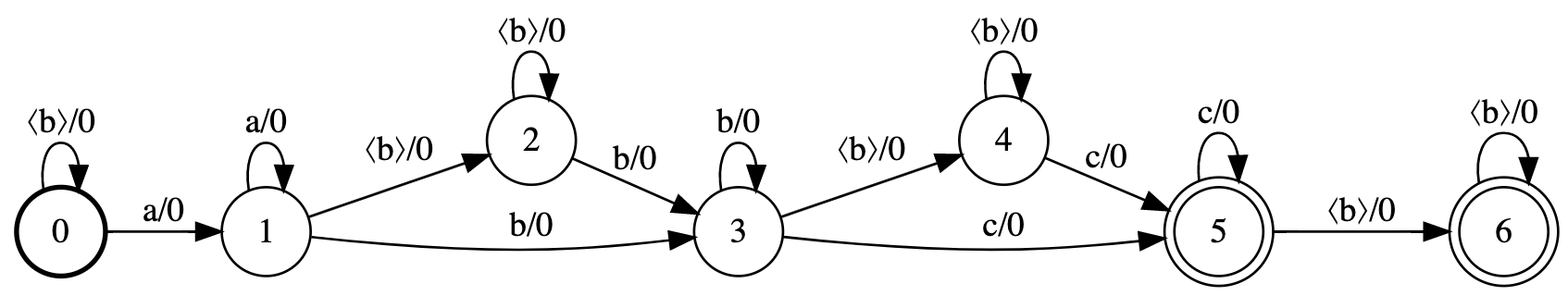}}
\end{floatrow}
\begin{floatrow}
\subfloat[Alignment Graph $\gA_{ctc} = \gY_{ctc} \circ \gE_{\mX} \label{fig:ctc_composed}$]{
    \includegraphics[width=0.5\linewidth]{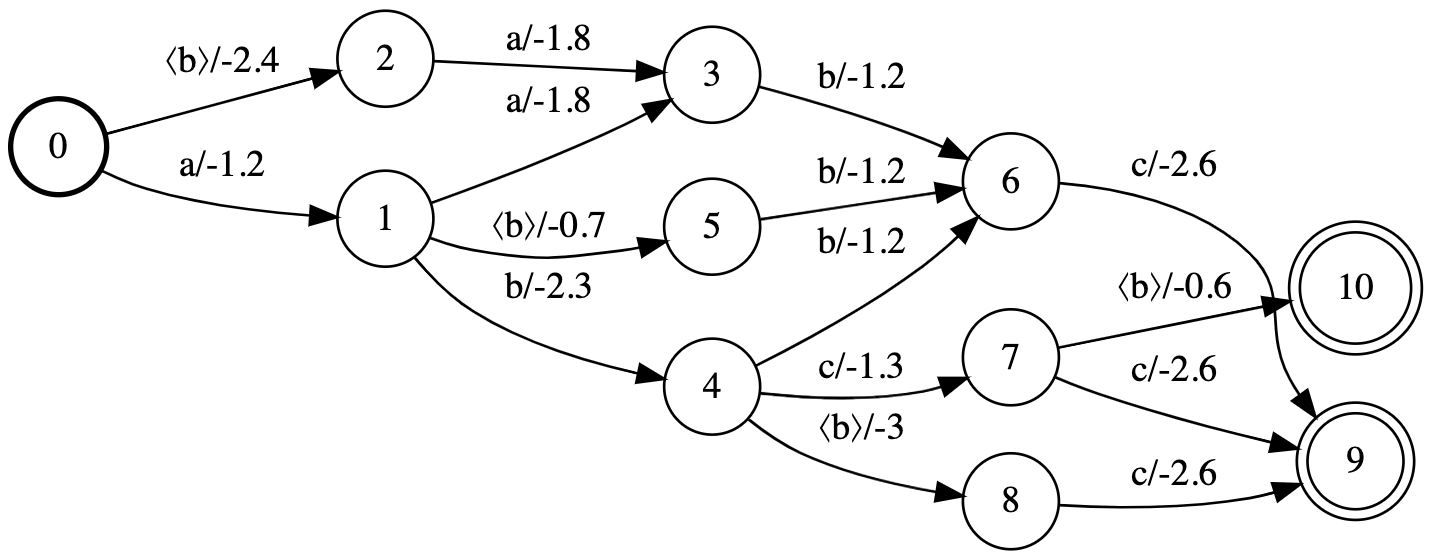}}
\subfloat[{Loss Graph $\gL_{ctc}=- \textit{Fwd}(\gA_{ctc}) \label{fig:ctc_loss}$}]{
\hspace{0.1\linewidth}
\raisebox{16pt}{\includegraphics[width=0.2\linewidth]{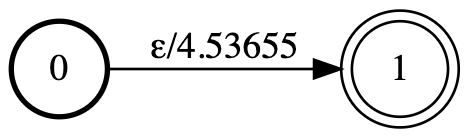}}
\hspace{0.1\linewidth}
}
\end{floatrow}
\vskip -0.15in
\caption{The sequence of steps involved in computing CTC loss using WFSTs. The arc label ``$p$/$w$" is a shorthand notation for ``$p$:$p$/$w$". $(a)$ is the emissions graph constructed from the log probabilities over the alphabet $\gA = \{a, b, c, d\}$ and blank symbol $\langle b \rangle$. The CTC label graph corresponding to the target sequence  ($a$, $b$, $c$) is shown in $(b)$. In $(c)$, we compose label graph with emission graph to get all valid paths which collapse to the target sequence. Finally, in $(d)$ we sum the probabilities of all of the valid paths (in log-space) and negate the result to yield the CTC loss.}
\label{fig:ctc_graphs}
\end{center}
\end{figure*}

The CTC loss can be computed purely with WFST operations as described in Figure~\ref{fig:ctc_graphs}. The composition of the emission graph (Figure~\ref{fig:ctc_emissions}) which has arc weights weights corresponding to $\log P(\pi_t \vert \rvx)$ and the label graph (Figure~\ref{fig:ctc_alignments}) is shown in Figure~\ref{fig:ctc_composed}. Each path from a start to a final state in Figure~\ref{fig:ctc_composed} is a valid CTC alignment. The negation of the forward score of the composed graph gives the CTC loss. These operations are differentiable and gradients can easily be propagated backward from the loss to a neural network to train the model.

Constructing the CTC loss from simpler WFST graphs also simplifies the implementation of variations of CTC. For example, the variations of \citet{moritz2021semi} only require changes to the label graph $\gY_{ctc}$. However, we when efficiency is a primary concern, custom GPU kernels can be faster than a generic implementation with WFSTs.

\subsection{Star Temporal Classification (STC)}
The label graph of CTC, $\gL_{ctc}$ (Figure~\ref{fig:ctc_alignments}), constructed from partial labels does not allow for the true target as a possibility. The STC algorithm addresses this problem by allowing for zero or more tokens from the alphabet between any two tokens in the partial label. Like CTC, an alignment between the input $\rvx$ and output $\rvy$ in STC is represented by a sequence of length $T$;  $\rvxpi =[\pi_1, \pi_2, ..., \pi_T]$ where $\pi_t \in \gA \cup \{\textit{blank}\}$ and $\rvy$ is a partial label of $\gB'(\pi)$. Unlike CTC, the STC collapse function $\gB'(\pi)$ only removes blank tokens. In other words, we do not allow self-loops on non \textit{blank} tokens in the STC label graph. This enables us to use a token insertion penalty as discussed in Section~\ref{sec:tip}.

\begin{figure}[ht]

\centering
\begin{floatrow}
\subfloat[Collapsing to $\langle s \rangle$ token \label{fig:star1}; w5 = logsumexp(w1,w2,w3,w4)]{
\begin{tabular}{ccc}
  {\includegraphics[width=0.39\textwidth]{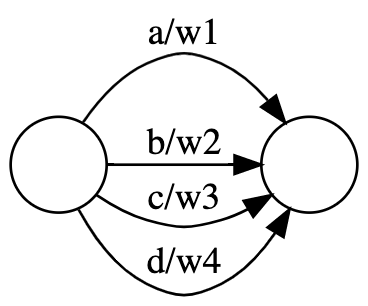}} & \raisebox{30pt}{$\rightarrow$} &  \raisebox{17pt}{\includegraphics[width=0.39\textwidth]{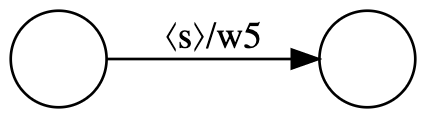}} \\
\end{tabular}
}
\end{floatrow}
\begin{floatrow}
\subfloat[{Collapsing to $\langle s \rangle \backslash a$ token \label{fig:star2}; w6 = logsumexp(w2,w3,w4)}]{
\begin{tabular}{ccc}
  {\includegraphics[width=0.39\textwidth]{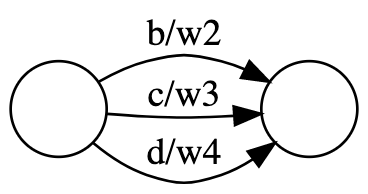}} & \raisebox{17pt}{$\rightarrow$} &  \raisebox{6pt}{\includegraphics[width=0.39\textwidth]{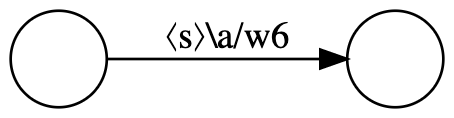}} \\
\end{tabular}
}
\end{floatrow}
\caption{Example showing the collapsing of tokens in alphabet $\gA = \{a, b, c, d\}$ to star token, $\langle s \rangle$}


 \label{fig:star_graphs}
\end{figure}

\begin{figure*}
\centering
\begin{floatrow}
\includegraphics[width=0.98\textwidth]{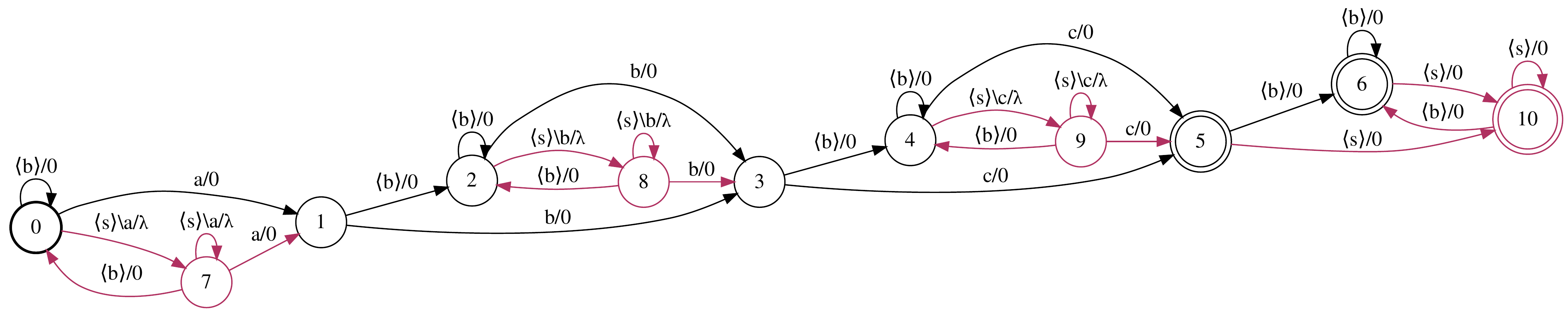}
\caption{STC Label Graph, $\gY_{stc}$ for the output sequence ($a$, $b$, $c$). In regex terms, this corresponds to ``\texttt{$[\:\hat{}\:a]^{*} a\:[\:\hat{}\:b]^{*} b\:[\:\hat{}\:c]^{*} c\:.^{*}$}". $\langle s \rangle, \langle b \rangle$ refer to \textit{star}, \textit{blank} tokens and $\langle s \rangle \backslash a$ is the relative complement of $a$ in $\langle s \rangle$. $\lambda (\leq 0)$ corresponds to token insertion penalty which has a regularization effect. The arc transitions marked in red show the changes from a selfless-CTC label graph.}
\label{fig:stc_emissions}
\end{floatrow}
\end{figure*}

To make the computation of the STC loss efficient, STC uses a special \textit{star} token, $\langle s \rangle$, which represents every token in the alphabet. It can be used to collapse the arcs between any two nodes in the WFST graph as shown in Figure~\ref{fig:star1}. We also define $\langle s \rangle \backslash t = \{y : y \in \gA; y \ne t\}$ which is the relative complement of $t$ in $\gA$. We use this in the label graph of STC, $\gL_{stc}$, to avoid counting the same alignment multiple times. The STC label graph, $\gL_{stc}$, for the partial label $(a, b, c)$ is shown in Figure~\ref{fig:stc_emissions}. The graph allows any alignment with zero or more tokens in between any two tokens of the given partial label sequence.

We also manually add the $\langle s \rangle$ and $\langle s \rangle \backslash t$ tokens to the emission graph $\gE_{\mX}$ with weights corresponding to their log-probabilities given by:
\begin{align}
\begin{split}\label{eq:prob_star}
   P(\langle s \rangle \vert \rvx) =& \sum_{y \in \gA} P(y \vert \rvx) \\
    P(\langle s \rangle \backslash t  \vert \rvx) =& \sum_{y \in \gA; y \ne t} P(y \vert \rvx).
\end{split}
\end{align}


\subsubsection{Token Insertion Penalty, $\lambda$}
\label{sec:tip}
We noticed that model training with STC does not work if we train directly with the STC label graph, even with $pDrop=0.1$. This is because the new paths introduced in STC from CTC (marked in red in Figure \ref{fig:stc_emissions}) would allow the model to produce a lot of possible output sequences for any given example and can confuse the model, especially during the early stages of training. To circumvent this problem, we use a parameter called token insertion penalty, $\lambda$ to add a penalty when using lot of new tokens. We use a exponential decay scheme to gradually reduce the penalty as the model starts training.  
\vskip -0.12in
\begin{equation}
\begin{array}{l}
    p_{t} = p_{max} + (p_0 - p_{max})  exp(- t / \tau) \\
    \lambda_{t} = ln(p_{t})
\end{array}
\end{equation}
\vskip -0.12in

where $p_0$, $p_{max}$, $\tau$ are hyperparameters and $\lambda_t$ denotes the token insertion penalty used in STC  for training step $t$. For choosing value of $\tau$, is is useful think in terms of half-life $t_{1/2} = \tau ln(2) $, which is the number of time steps taken for $p_t$ to reach $(p_0 + p_{max})/2$ 


\subsubsection{Implementation Details}
\label{sec:impl}
We use the CPU-based WFST algorithms from GTN~\cite{hannun2020differentiable} to implement the STC criterion. Multiple threads are used to compute STC loss in parallel for all of the examples in a batch. Since the neural network model is run on GPU, the emissions needed by STC must be copied to the CPU, and the STC gradients must be copied back to GPU. To reduce the amount of data transfer between the CPU and the GPU, we transfer values corresponding only to the tokens present in the partial labels and the \textit{star} tokens. This works since the gradients corresponding to all other tokens are zero. 

\begin{figure}[ht]
\centering
\begin{floatrow}
\includegraphics[width=0.98\textwidth]{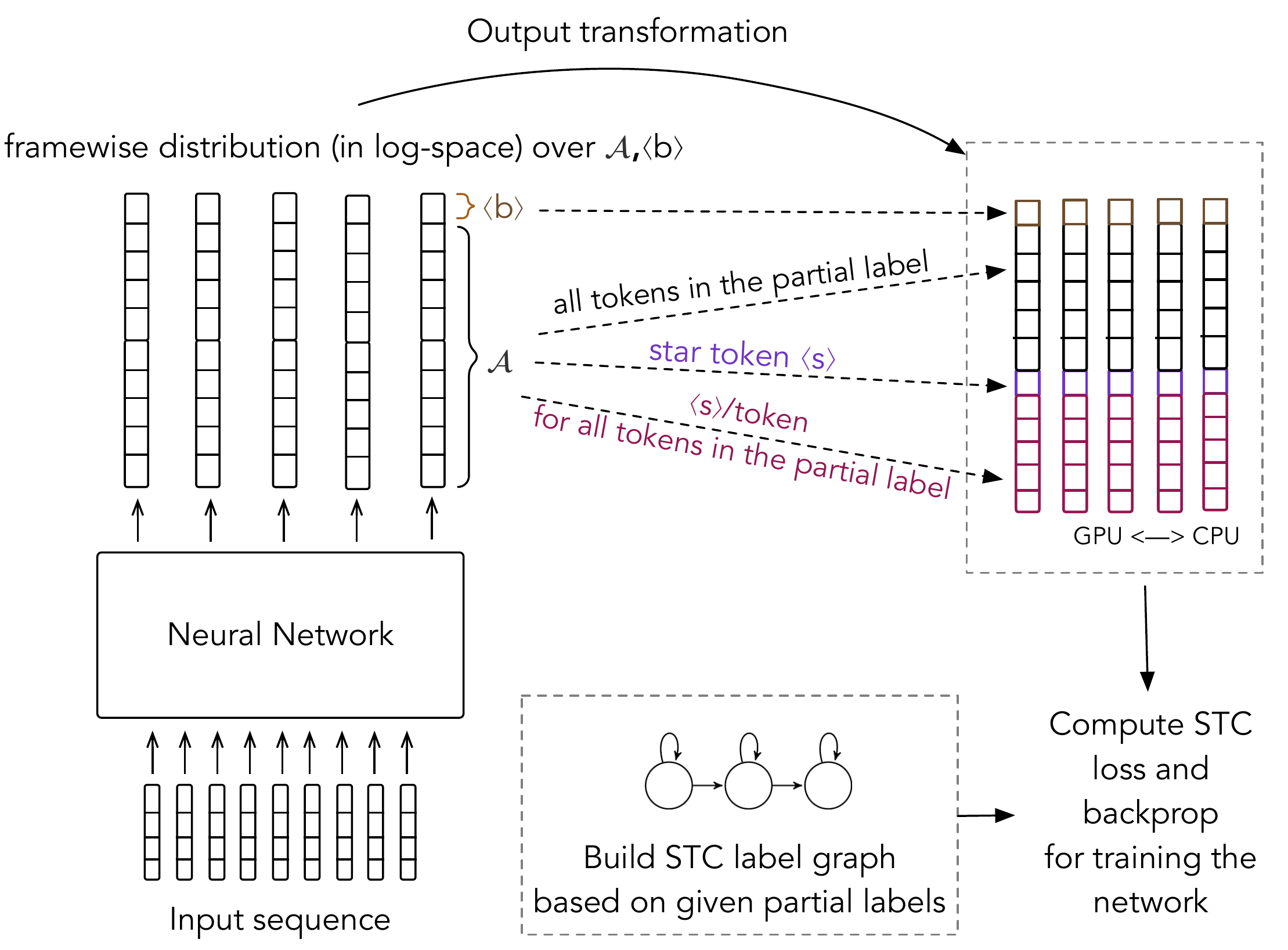}
\caption{The STC Training Pipeline}
\label{fig:stc_arch}
\end{floatrow}
\vskip -0.1in 
\end{figure}

Figure~\ref{fig:stc_arch} shows an overview of the STC training pipeline. The input sequence is typically passed through a neural network on GPU to produce a frame wise distribution (in log-space)  over $\gA \cup \{\textit{blank}\}$. The output number of frames depend on the size of the input sequence, as well as the amount of padding and the stride of the neural network model architecture. We then compute the log-probability of star token, $\langle s \rangle$ and $\langle s \rangle \backslash t$ tokens  $ \forall t \in \rvy $ using Equation~\ref{eq:prob_star}. We construct the emission graph $\gE_{\mX}$ with weights as log-probabilities and the label graph $\gL_{stc}$ on CPU. Using the WFST operations shown in Figure~\ref{fig:ctc_graphs}, we compute STC loss and gradients of loss with respect to arc weights (log probability output) of $\gE_{\mX}$. These gradients are copied back to GPU and the gradients of parameters in neural network are computed using backpropagation. Then the model can be trained using standard gradient descent methods. 




\section{Experimental Setup}



\subsection{Automatic Speech Recognition (ASR)} 
We use LibriSpeech \cite{panayotov2015librispeech} dataset, containing 960 hours of training audio with paired transcriptions for our speech recognition experiments. The standard LibriSpeech validation sets (\textit{dev-clean} and \textit{dev-other}) are used to tune all hyperparameters, as well as to select the best models. We report final word error rate (WER) performance on the test sets (\textit{test-clean} and \textit{test-other}). 

We keep the original 16kHz sampling rate and compute log-mel filterbanks with 80 coefficients for a 25ms sliding window, strided
by 10ms. All features are normalized to have zero mean and unit variance per input sequence before feeding them into the acoustic model. We use SpecAugment~\cite{park_spec} as the data augmentation method for all ASR experiments.

We use the top 50K words (sorted by occurance frequency) from the official language model (LM) training data provided with LibriSpeech as the alphabet for training the models. These 50K words cover 99.04\% of all the word occurances in the LM training data.  The model architectures and the STC loss are implemented with the ASR application~\cite{pratapw2l} of the flashlight\footnote{\url{https://github.com/flashlight/flashlight}} machine-learning framework and the C++ API of GTN.

\subsubsection{Label Generation for Weakly Supervised Setup} 
\label{lbl_gen}
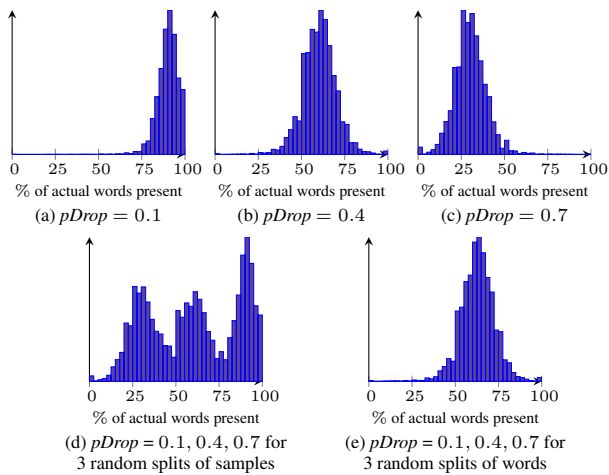
\begin{figure}[ht]
\pgfplotsset{
    ybar, 
    title style={font=\scriptsize},
    label style={font=\tiny, yshift=3pt},
    xticklabel style={font=\tiny, yshift=2pt, xshift=1pt},
    xlabel={$\%$ of actual words present},
    ylabel={},
    ymin=0,
    ytick=\empty,
    xtick={0, 25, 50, 75, 100},
    axis x line=bottom,
    axis y line=left,
    title style={at={(0.5,-0.42)},anchor=north,yshift=-0.1},
}
\pgfplotsset{/pgfplots/group/.cd,
    horizontal sep=0.4cm,
}
\begin{minipage}{\textwidth}
\begin{tikzpicture}
\begin{groupplot}[group style={group size=3 by 1},height=3.5cm,width=0.47\textwidth]

\nextgroupplot[title={(a) \textit{pDrop} $=0.1$}]
\addplot+[ybar interval,    fill=Violet,mark=no] plot coordinates {  (0.0,16.0) (2.5,0.0) (5.0,0.0) (7.5,0.0) (10.0,0.0) (12.5,0.0) (15.0,0.0) (17.5,0.0) (20.0,2.0) (22.5,0.0) (25.0,3.0) (27.5,0.0) (30.0,0.0) (32.5,12.0) (35.0,0.0) (37.5,1.0) (40.0,15.0) (42.5,5.0) (45.0,4.0) (47.5,0.0) (50.0,122.0) (52.5,9.0) (55.0,85.0) (57.5,15.0) (60.0,178.0) (62.5,147.0) (65.0,594.0) (67.5,260.0) (70.0,1084.0) (72.5,1038.0) (75.0,3944.0) (77.5,5077.0) (80.0,13572.0) (82.5,21840.0) (85.0,32479.0) (87.5,47369.0) (90.0,54552.0) (92.5,45402.0) (95.0,28898.0) (97.5,24518.0) (100, 0) };

\nextgroupplot[title={(b) \textit{pDrop} $=0.4$}]
\addplot+[ybar interval,fill=Violet,mark=no,title=$pDrop=0.4$] plot coordinates {  (0.0,150.0) (2.5,0.0) (5.0,0.0) (7.5,1.0) (10.0,18.0) (12.5,54.0) (15.0,84.0) (17.5,19.0) (20.0,216.0) (22.5,40.0) (25.0,426.0) (27.5,307.0) (30.0,405.0) (32.5,1156.0) (35.0,996.0) (37.5,1840.0) (40.0,3885.0) (42.5,5708.0) (45.0,8696.0) (47.5,7715.0) (50.0,23270.0) (52.5,23495.0) (55.0,28831.0) (57.5,28214.0) (60.0,33270.0) (62.5,31135.0) (65.0,25088.0) (67.5,18188.0) (70.0,14640.0) (72.5,7521.0) (75.0,6916.0) (77.5,2684.0) (80.0,2680.0) (82.5,1203.0) (85.0,651.0) (87.5,599.0) (90.0,318.0) (92.5,54.0) (95.0,3.0) (97.5,765.0) (100, 0)};

\nextgroupplot[title={(c) \textit{pDrop} $=0.7$}]
\addplot+[ybar interval,fill=Violet,mark=no] plot coordinates {    (0.0,1783.0) (2.5,80.0) (5.0,539.0) (7.5,1158.0) (10.0,2453.0) (12.5,5218.0) (15.0,8102.0) (17.5,10785.0) (20.0,21434.0) (22.5,21580.0) (25.0,35600.0) (27.5,32558.0) (30.0,34762.0) (32.5,29263.0) (35.0,23076.0) (37.5,17298.0) (40.0,14257.0) (42.5,8033.0) (45.0,4876.0) (47.5,1506.0) (50.0,3482.0) (52.5,908.0) (55.0,837.0) (57.5,207.0) (60.0,474.0) (62.5,250.0) (65.0,341.0) (67.5,18.0) (70.0,101.0) (72.5,12.0) (75.0,77.0) (77.5,11.0) (80.0,52.0) (82.5,23.0) (85.0,6.0) (87.5,9.0) (90.0,0.0) (92.5,0.0) (95.0,0.0) (97.5,72.0) (100, 0)};
  \end{groupplot}
\end{tikzpicture}
\end{minipage}

\begin{minipage}{\textwidth}
\begin{tikzpicture}
\begin{groupplot}[group style={group size=3 by 1},height=3.5cm,width=0.47\textwidth]
\nextgroupplot[title={}, hide axis]
\nextgroupplot[xshift=-0.6in, align=center, title={{(d) \textit{pDrop} = $0.1,0.4,0.7$ for \\ 3 random splits of samples}}]
\addplot+[ybar interval,fill=Violet,mark=no] plot coordinates {(0.0,685.0) (2.5,20.0) (5.0,200.0) (7.5,362.0) (10.0,762.0) (12.5,1826.0) (15.0,2746.0) (17.5,3659.0) (20.0,7387.0) (22.5,7037.0) (25.0,12017.0) (27.5,11172.0) (30.0,11960.0) (32.5,10049.0) (35.0,7852.0) (37.5,6341.0) (40.0,5989.0) (42.5,4532.0) (45.0,4463.0) (47.5,3046.0) (50.0,8963.0) (52.5,8114.0) (55.0,9930.0) (57.5,9504.0) (60.0,11347.0) (62.5,10394.0) (65.0,8694.0) (67.5,6300.0) (70.0,5205.0) (72.5,2879.0) (75.0,3754.0) (77.5,2732.0) (80.0,5530.0) (82.5,7686.0) (85.0,10935.0) (87.5,15988.0) (90.0,18274.0) (92.5,14945.0) (95.0,9565.0) (97.5,8397.0) (100, 0)};

\nextgroupplot[xshift=0.4in, align=center, title={{(e) \textit{pDrop} = $0.1,0.4,0.7$ for \\ 3 random splits of words}}]
\addplot+[ybar interval,fill=Violet,mark=no,title=$pDrop=0.4$] plot coordinates {   (0.0,141.0) (2.5,0.0) (5.0,0.0) (7.5,3.0) (10.0,11.0) (12.5,51.0) (15.0,68.0) (17.5,9.0) (20.0,179.0) (22.5,18.0) (25.0,309.0) (27.5,197.0) (30.0,196.0) (32.5,789.0) (35.0,570.0) (37.5,1060.0) (40.0,2437.0) (42.5,3436.0) (45.0,5305.0) (47.5,4686.0) (50.0,16426.0) (52.5,17324.0) (55.0,23302.0) (57.5,24829.0) (60.0,32538.0) (62.5,33716.0) (65.0,30051.0) (67.5,24273.0) (70.0,20862.0) (72.5,11982.0) (75.0,11603.0) (77.5,5014.0) (80.0,4331.0) (82.5,2035.0) (85.0,1114.0) (87.5,858.0) (90.0,500.0) (92.5,108.0) (95.0,13.0) (97.5,897.0) (100, 0)};
\end{groupplot}
\end{tikzpicture}
\end{minipage}

\caption{Histograms of percentage of transcript words retained in each training sample in LibriSpeech for different ways of generating weakly supervised data using $pDrop$. }
\label{fig:hist_dist}
\end{figure}

As LibriSpeech dataset consists of fully labeled data, we need to drop the labels manually to simulate the weakly supervised setup. We quantify this using a parameter $pDrop \in [0, 1]$, which denotes the probability of dropping a word from the transcript label. It can be seen that higher value of  $pDrop$ corresponds to higher number of missing words in the transcript and vice-versa. Figures \ref{fig:hist_dist}a-c show the histogram of percentage of the labeled words present in each
training sample in LibriSpeech for different values of $pDrop$ which closely follows a normal distribution.  
Also, from the definition of $pDrop$, it naturally follows that $pDrop=0$ corresponds to supervised training setup while $pDrop=1$ corresponds to unsupervised training.

To make sure our method is robust to different types of partial labels, we also test our method on two other ways of generating the data : \\
1. Randomly split the training set into $p$ parts  and use a different $pDrop$ for each split (Figure~\ref{fig:hist_dist}d) \\
2. Randomly split the words into $p$ parts, use a different $pDrop$ for each split (Figure~\ref{fig:hist_dist}e) 

For all the new partially labeled datasets created, we prune samples which have empty transcriptions before using them for training.  

\subsubsection{Model architecture}
The acoustic model (AM) architecture is composed
of a convolutional frontend (1-D convolution with kernel-width
15 and stride 8 followed by GLU activation) followed by 36 $\times$ 4-
heads Transformer blocks \cite{vaswani2017attention} with relative positional embedding. The self-attention dimension is 384 and the feed-forward network (FFN) dimension is 3072 in each Transformer block. The output of the final Transformer block if followed by a Linear layer with output dimension of 580 and a letter-to-word encoder (see Appendix~\ref{apx:l2w}) to the output classes (word vocabulary + \textit{blank}). For all Transformer layers, we use dropout on the self-attention and on the FFN, and layer drop~\cite{fan2019reducing}, dropping entire layers at the FFN level. We apply LogSoftmax operation on each output frame to produce a probability distribution (in log-space) over output classes. The model  consists of 70 million parameters and we use 32 $\times$ Nvidia 32GB V100 GPUs for training. 

\subsubsection{Decoding}
\label{sec:decoding}
\textbf{{Beam-Search decoder}}:
In our experiments, we use a beam-search decoder following~\cite{collobert2020word} which leverages a $n$-gram language model to decrease the word error rate. The beam-search decoder outputs a transcription $\hat{\rvy}$ that maximizes the following objective 
\begin{align}
    \log P(\hat{\rvy} \vert \rvx) + \alpha \log P_{LM}(\hat{\rvy}) + \beta \lvert \hat{\rvy} \rvert 
\end{align}

where $\log P_{LM}(.)$ is the log-likelihood of the language model, $\alpha$ is the weight of the language model, $\beta$ is a word insertion weight and $ \lvert \hat{\rvy} \rvert $ is the transcription length in words. The hyperparameters $\alpha$ and $\beta$ are optimized on the validation set. 

We use 5-gram LM trained on the official LM training data provided with LibriSpeech for beam-search decoding.

\textbf{{Rescoring}}: To further decrease the word error rate, we use the top-$K$ hypothesis from beam search decoding and perform rescoring~\cite{synnaeve2020endtoend} with a Transformer LM to reorder the hypotheses according to the following score:
\begin{align}
    \log P(\hat{\rvy} \vert \rvx) + \alpha \log P_{LM'}(\hat{\rvy}) + \beta \lvert \hat{\rvy} \rvert 
\end{align}

where $\log P_{LM'}(.)$ is the log-likelihood of the Transformer LM, $\alpha$, $\beta$ are the hyperparameters as described above and $ \lvert \hat{\rvy} \rvert $ is the transcription length in characters. We use the pre-trained Transformer LM on LibriSpeech from~\cite{likhomanenko21_interspeech} which has a perplexity of 50 on dev-other transcripts. In this work, top 512 hypothesis from beam search decoding are used as candidates for rescoring.

\subsubsection{Pseudo-labeling}
To further achieve better word error rate (WER), we  generate pseudo labels (PLs) on the training set using the word-based model trained with STC and use these PLs to train a letter-based model, via a regular CTC approach. The PLs are generated using rescoring as described in Section~\ref{sec:decoding} on the training set using the hyperparameters optimized on validation set.  We use the training recipe used by \cite{pratap2021word} for training the letter-based CTC models which uses a Transformer-based acoustic model consisting of 270 million parameters and uses 64 $\times$ Nvidia 32GB V100 GPUs for training.  

\begin{table*}[ht!]
\caption{WER comparison on LibriSpeech dev and test sets for various partial label generation settings. For each setting, we report the results for training with CTC (control experiment), STC on the partially labeled data and training with CTC using PLs generated from the trained STC model. Whenever possible, for each of these experiments, we report WER with a greedy decoding and no LM (top row), with 5-gram LM beam-search decoding (middle row) and with additional second-pass rescoring by Transformer LM (below row). We also include state-of-the-art results on LibriSpeech using wordpieces, letters and words as output tokens in the top section of the table.}
\label{ls-wer1}
\begin{center}
\begin{scriptsize}
\begin{sc}
\begin{tabular}{cccccccc}
\toprule
 Weak label & \multirow{2}{*}{Method} & \multirow{2}{*}{Criterion} & Model Stride/ & Output &  \multirow{2}{*}{LM}  &  {Dev WER} & {Test WER}  \\
gen. Strategy & & & Parameters & tokens & &  clean/other & clean/other \\
\midrule
 & Conformer  & \multirow{2}{*}{seq2seq} & \multirow{2}{*}{4/119M}  & 1K Word & - & 1.9/4.4 & 2.1/4.3\\
 & \cite{gulati20_interspeech} & & & pieces & LSTM & & 1.9/3.9\\

\rule{0pt}{3ex}     & \multirow{2}{*}{Transformer}  &\multirow{3}{*}{CTC} & \multirow{3}{*}{3/270M} & \multirow{3}{*}{letters} &  - & 2.5/5.9 & 2.7/6.1  \\
 &  & & & & word 5-gram & 1.9/4.7 & 2.4/5.3 \\
 & \cite{pratap2021word} & & & & Tr. Rescoring & 1.6/4.0 & 2.1/4.5 \\ 
 \rule{0pt}{3ex}  &  {Transformer}  & \multirow{2}{*}{seq2seq} &   \multirow{2}{*}{8/$\sim$ 300M} & \multirow{2}{*}{Words} &   - & 2.7/6.5 & 2.9/6.7 \\
   & \tiny{\cite{collobert2020word}} & & & & word 4-gram & 2.5/6.0 & 3.0/6.3 \\
\midrule
  &  \multirow{3}{*}{Transformer} & \multirow{3}{*}{CTC} & \multirow{3}{*}{3/270M} & \multirow{3}{*}{letters} &  - & 6.0/11.1 & 6.2/11.0 \\
  & & & & &  Word 5-gram & 4.8/8.7 & 4.9/8.8  \\  
  & & & & &  Tr. Rescoring & 4.4/7.9 & 4.5/8.1  \\ \cdashline{2-8}
  &  \multirow{3}{*}{Transformer}  & \multirow{3}{*}{STC} & \multirow{3}{*}{8/70M} & \multirow{3}{*}{Words} &  \rule{0pt}{2.5ex} - & 3.9/8.9 & 4.1/9.0 \\ 
{pDrop = 0.1} & & & & &  Word 5-gram & 3.9/8.5 & 4.1/8.7  \\
& & & & &  Tr. Rescoring  & 3.0/6.6 & 3.3/6.9 \\
& \multirow{3}{*}{$+$ Pseudo Labeling}  & \multirow{3}{*}{CTC} & \multirow{3}{*}{3/270M} & \multirow{3}{*}{letters} &  \rule{0pt}{2.5ex} - & 2.9/6.0 & 3.1/6.3 \\
  & & & & &  Word 5-gram & 2.6/5.2 & 2.9/5.6   \\  
  & & & & &  Tr. Rescoring &  2.4/4.7 & 2.7/5.2  \\
\midrule
  &  \multirow{1}{*}{Transformer} & \multirow{1}{*}{CTC} & \multirow{1}{*}{3/270M} & \multirow{1}{*}{letters} &   - & 49.5/57.8 & 48.2/56.9 \\ \cdashline{2-8}
  &  \multirow{3}{*}{Transformer}  & \multirow{3}{*}{STC} & \multirow{3}{*}{8/70M} & \multirow{3}{*}{Words} &  \rule{0pt}{2.5ex}- & 4.5/10.0 & 4.5/10.2 \\ 
{pDrop = 0.4} & & & & &  Word 5-gram & 4.6/9.6 & 4.6/9.9  \\
& & & & &  Tr. Rescoring  & 3.4/7.3 & 3.6/7.7\\
& \multirow{3}{*}{$+$ Pseudo Labeling}  & \multirow{3}{*}{CTC} & \multirow{3}{*}{3/270M} & \multirow{3}{*}{letters} &  \rule{0pt}{2.5ex} - &  3.1/6.3 & 3.3/6.6 \\
  & & & & &  Word 5-gram & 2.8/5.4 & 3.1/5.9  \\  
  & & & & &  Tr. Rescoring &  2.6/4.9  & 2.9/5.4\\
\midrule
    &  \multirow{1}{*}{Transformer} & \multirow{1}{*}{CTC} & \multirow{1}{*}{3/270M} & \multirow{1}{*}{letters} &  - & 100/100 & 100/100 \\ \cdashline{2-8}
  &  \multirow{3}{*}{Transformer}  & \multirow{3}{*}{STC} & \multirow{3}{*}{8/70M} & \multirow{3}{*}{Words} & \rule{0pt}{2.5ex} - & 6.8/14.7  & 6.9/15.1\\ 
{pDrop = 0.7} & & & & &  Word 5-gram &  7.0/14.2 & 7.0/14.8 \\
& & & & &  Tr. Rescoring  & 5.0/10.7 &  5.1/11.1 \\
& \multirow{3}{*}{$+$ Pseudo Labeling}  & \multirow{3}{*}{CTC} & \multirow{3}{*}{3/270M} & \multirow{3}{*}{letters} &  \rule{0pt}{2.5ex} - &  3.5/7.4 & 3.9/7.9 \\
  & & & & &  Word 5-gram & 3.2/6.7 & 3.5/7.0  \\  
  & & & & &  Tr. Rescoring &  2.9/5.8 & 3.1/6.2 \\

\midrule
  &  \multirow{1}{*}{Transformer} & \multirow{1}{*}{CTC} & \multirow{1}{*}{3/270M} & \multirow{1}{*}{letters} &  - & 58.9/64.2 & 58.6/63.9 \\ \cdashline{2-8}
Split all samples  &  \multirow{3}{*}{Transformer}  & \multirow{3}{*}{STC} & \multirow{3}{*}{8/70M} & \multirow{3}{*}{Words} & \rule{0pt}{2.5ex} - & 4.8/10.7  & 4.9/11.1 \\ 
into 3 parts & & & & &  Word 5-gram & 5.0/10.6 & 5.1/11.0  \\
 randomly; & & & & &  Tr. Rescoring  & 3.5/7.9 & 3.7/8.2 \\
Assign pDrop=& \multirow{3}{*}{$+$ Pseudo Labeling}  & \multirow{3}{*}{CTC} & \multirow{3}{*}{3/270M} & \multirow{3}{*}{letters} &  \rule{0pt}{2.5ex} -  & 3.1/6.5 & 3.3/6.7 \\
 0.1.0.4,0.7 for & & & & &  Word 5-gram & 2.8/5.7 & 3.0/5.9  \\  
the splits  & & & & &  Tr. Rescoring & 2.7/5.0 & 2.8/5.4  \\
\midrule
    &  \multirow{1}{*}{Transformer} & \multirow{1}{*}{CTC} & \multirow{1}{*}{3/270M} & \multirow{1}{*}{letters} &  - & 45.1/49.1 & 45.6/49.2 \\ \cdashline{2-8}
Split all words   &  \multirow{3}{*}{Transformer}  & \multirow{3}{*}{STC} & \multirow{3}{*}{8/70M} & \multirow{3}{*}{Words} & \rule{0pt}{2.5ex} - & 5.8/12.5  & 5.8/13.0 \\ 
into 3 parts & & & & &  Word 5-gram & 6.4/11.8 & 6.6/12.2   \\
 randomly; & & & & &  Tr. Rescoring  & 4.0/8.3 & 4.1/8.7  \\
Assign pDrop= & \multirow{3}{*}{$+$ Pseudo Labeling}  & \multirow{3}{*}{CTC} & \multirow{3}{*}{3/270M} &  \multirow{3}{*}{letters} & \rule{0pt}{2.5ex} - & 3.2/6.7 & 3.4/6.9 \\
0.1.0.4,0.7 for  & & & & &  Word 5-gram &  2.8/5.4 & 3.1/5.9  \\  
the splits  & & & & &  Tr. Rescoring & 2.6/5.1 & 2.9/5.6  \\

\bottomrule
\end{tabular}
\end{sc}
\end{scriptsize}
\end{center}
\end{table*}

\subsection{Handwriting Recognition (HWR)} 
We test our approach on IAM Handwriting database \cite{marti2002iam}, which is a widely used benchmark for handwriting recognition. The dataset
contains 78 different characters and a white-space symbol, $\mid$. We use Aachen data splits\footnote{\footnotesize \url{https://www.openslr.org/56}} to divide the dataset into three subsets: 6,482 lines for training, 976 lines for validation and 2,915 lines for testing. 

\begin{figure}[ht]
\begin{center}
\centerline{\includegraphics[width=0.98\textwidth]{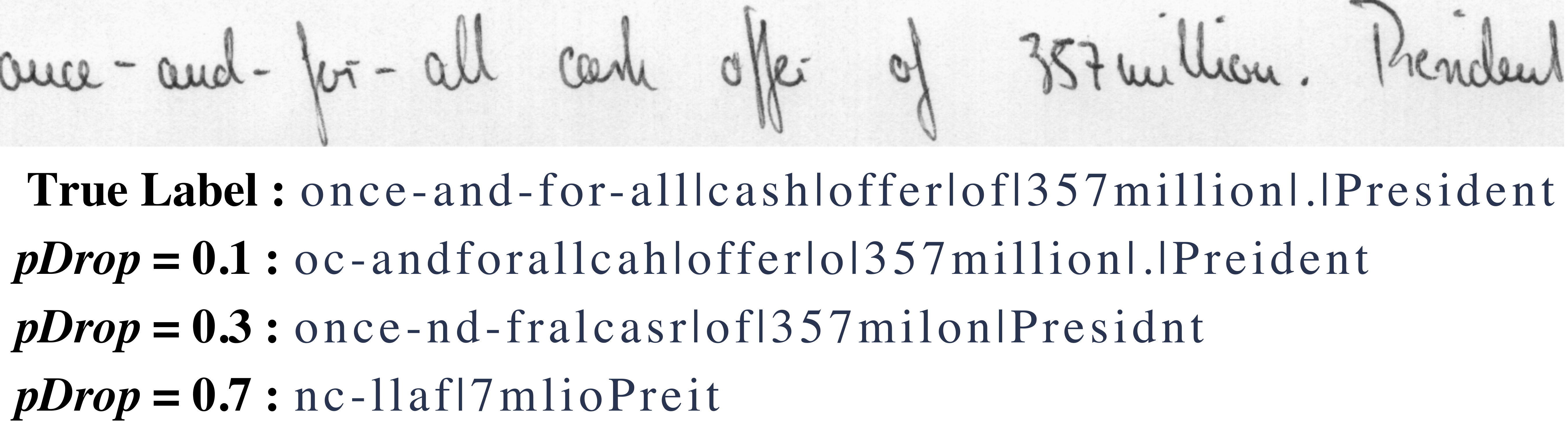}}
\vspace{-0.4in}
\caption{A training example from IAM along with weakly supervised labels generated for different $pDrop$ values.}
\label{fig:iam_data}
\end{center}
\end{figure}

To create weakly supervised labels, we use the same methodology using $pDrop$ as described in \ref{lbl_gen}. However, we drop the labels at character-level instead of word-level. Figure \ref{fig:iam_data} shows a training example from IAM along with weakly supervised labels generated for different $pDrop$ values. 

For training the handwriting recognition models, we have used the open-source code\footnote{\footnotesize \url{ https://github.com/IntuitionMachines/OrigamiNet}} based on a prior work from  \citet{yousef2020origaminet} and adapted it for our use case. It uses depthwise separable convolutions as the main computational block and the model consists of about $39$ million parameters. We use 8 $\times$ Nvidia V100 32GB GPUs for training the models.  


Additional implementation and training details (models, tokens, optimization, other hyperparameters and settings) for the ASR and HWR experiments are in the Appendix~\ref{apx:detail}. 
\section{Results}

\subsection{Automatic Speech Recognition} 
In Table \ref{ls-wer1}, we compare the WER performance of 
STC models for a fixed value of $pDrop = 0.1, 0.4, 0.7$ and also when $pDrop$ is dependent on the sample or the word. We also compare these results with the models trained directly with CTC on the partial labels. We can see that STC performs better than CTC in all the settings. We observe that performing beam search decoding and an additional second-pass rescoring with a Transformer LM can reduce the WER of the models significantly.

 We also see that pseudo labeling the training set using STC model trained on partial labels and then training a letter-based CTC model can further improve WER performance. We can get competitive results compared with the fully supervised results. 



\subsection{Handwriting Recognition} 
In Table~\ref{iam-cer}, we compare the CER performance of CTC and STC for various values of $pDrop$ with greedy decoding. We can see that using STC clearly gives better performance over CTC trained models for $pDrop = 0.1, 0.3, 0.5, 0.7$. The performance gap between the supervised baseline and models trained with partial labels using STC is larger compared to ASR experiments. This is because we did not use LM decoding or pseudo-labeling steps which should help in reducing CER.

Hyperparameters used for token insertion penalty, $\lambda$ to train STC models for ASR, HWR experiments are in Appendix~\ref{apx:tip}. 

\begin{table}[ht]
\caption{CER performance comparison of STC and CTC with greedy decoding on IAM dataset for different $pDrop$ values. A comparison with selected works on IAM dataset using the ``same" training set is shown at the top.}
\label{iam-cer}
\begin{center}
\begin{scriptsize}
\begin{sc}
\begin{tabular}{@{\hskip 0in}clcc}
\toprule
pDrop & Method &  {Dev CER} & {Test CER}  \\
\midrule
  & \multirow{2}{*}{\parbox{3cm}{Transformer \\ \cite{kang2020pay}}}  &   & 4.7 \\ \\
  & \multirow{2}{*}{\parbox{3cm}{CNN+CTC \\ \cite{yousef2020accurate}}}  &   & 4.9 \\ \\
&  \multirow{2}{*}{\parbox{3cm}{LSTM w/Attn \\ \cite{michael2019evaluating}}}  &  3.2 & 5.5 \\ \\  

\midrule
\midrule
&  \multirow{2}{*}{\parbox{3cm}{CNN + CTC \\ (Our Baseline)}}  &  3.7 & 5.4 \\
\\
\midrule
\multirow{2}{*}{0.1} & CNN + CTC &  5.4 & 7.7 \\
 & CNN + STC & 5.0  & 7.2  \\
\midrule
\multirow{2}{*}{0.3} & CNN + CTC &   8.7  & 11.6 \\
 & CNN + STC & 5.6  & 8.1  \\
\midrule
\multirow{2}{*}{0.5} & CNN + CTC &  48.2 & 53.6\\
 & CNN + STC &  10.0 & 13.5  \\
 \midrule
\multirow{2}{*}{0.7} & CNN + CTC &  77.3  & 78.5\\
 & CNN + STC & 22.7  & 26.7  \\
\bottomrule
\end{tabular}
\end{sc}
\end{scriptsize}
\end{center}
\vspace{-0.1in} 
\end{table}


\subsection{Runtime Performance of STC}
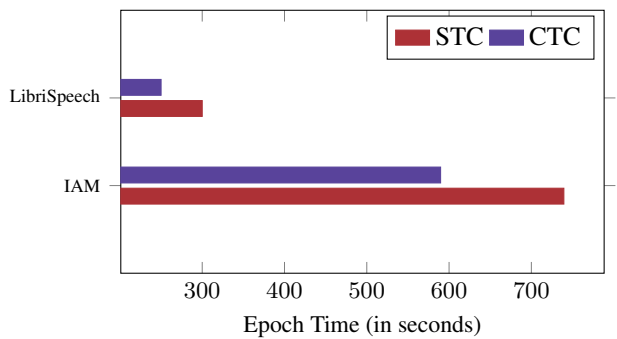
\begin{figure}[ht]
\centering
\begin{tikzpicture}
\begin{axis}[
height=2in,
width=0.97\textwidth,
legend columns=-1, 
xlabel={Epoch Time (in seconds)},
ytick={1, 2},
ymin=0,
ymax=3,
yticklabels={{IAM}, LibriSpeech},
xbar,
nodes near coords align={horizontal},
bar width=6pt,
xlabel style = {font=\footnotesize},
yticklabel style = {font=\scriptsize},
xticklabel style = {font=\footnotesize},
legend image code/.code={
        \draw [#1] (0cm,-0.1cm) rectangle (0.45cm,0.1cm); },
]
\addplot [Maroon,fill=Maroon] coordinates {
(300, 2 ) (739.5, 1) 
};
\addplot [Violet,fill=Violet] coordinates {
 (250,2  ) (589.7, 1) 
};

\legend{STC, CTC}
\end{axis}
\end{tikzpicture}
\caption{Epoch time performance of STC, CTC on LibriSpeech and IAM datasets for the models we have used.}
\vspace{-0.2in}
\label{fig:perf1}
\end{figure}

In Figure~\ref{fig:perf1}, we compare the epoch time of STC, CTC models on LibriSpeech and IAM datasets. For a fair comparison, we use the same model that is used to report STC results for each dataset. The experiments on LibriSpeech, IAM are run on 32, 8 Nvidia 32GB V100 GPUs and uses C++, Python APIs of GTN respectively.

We were able to run an epoch on LibriSpeech in about 300 minutes for the STC model and in about 250 seconds for the CTC model. Optimizing the data transfer between CPU and GPU by only moving the tokens which are present in the current training sample (Section~\ref{sec:impl}) is a crucial step in achieving good performance. Without this optimization, it would take about 5400 seconds per epoch on LibriSpeech as the alphabet size, $\lvert \gA \rvert$ of 50000 is very high. 

For IAM experiments, we did not perform the data transfer optimization for STC as $\lvert \gA \rvert$ is only 79. It can be see that the epoch time of STC is about 25\% more compared to CTC.   

\vspace{-0.2in}
\section{Conclusion}

In a variety of sequence labeling tasks (ASR, HWR), we show that STC enables training models with partially labeled data and can give strong performance. Weakly supervised data can be collected in semi-automatic ways and can alleviate labeling of full training data, thus reducing production costs. We also show that using WFSTs with a differentiable framework is flexible and powerful tool for researchers to solve a completely new set of problems, like the example we have shown in this paper. A direct potential extension is to study the application of STC to noisy labels.


\bibliography{references}
\bibliographystyle{icml2022}

\newpage
\appendix
\onecolumn
\section{Additional Implementation Details}
\label{apx:detail}
\subsection{Automatic Speech Recognition}

For training the word-based STC models, we use the Adagrad optimizer~\cite{duchi11a_adagrad} with a learning rate warmup scheme that increases linearly from $0$ to $0.02$ in 16000 training steps for all the experiments. We halve the learning rate initially after 400 epochs and every 200 epochs after that. All models are trained with dynamic batching with a batch size of 240 audio sec per GPU. We use SpecAugment~\cite{park_spec} with two frequency masks, and ten time masks with maximum time mask ratio of $p$ = 0.1, the maximum frequency bands masked by one frequency mask is 30, and the maximum frames masked by the time mask is 50; time warping is not used. SpecAugment is turned on only after 32000 training steps are finished. Dropout and layer dropout values of 0.05 is used in the AM.

For training the letter-based CTC models,  we use the same Transformer-based encoder consisting of 270M parameters from~\cite{synnaeve2020endtoend} for the AM: the encoder of our acoustic models is composed of a convolutional frontend (1-D convolution with kernel-width~7 and stride~3 followed by GLU activation) followed by 36 4-heads Transformer blocks~\cite{vaswani2017attention}. The self-attention dimension is $768$ and the feed-forward network (FFN) dimension is $3072$ in each Transformer block.
The output of the encoder 
is followed by a linear layer to the output classes. 
We use dropout after the convolution layer. For all Transformer layers, we use dropout of $0.2$ on the self-attention and on the FFN, and layer drop~\cite{fan2019reducing} value of $0.2$. Token set for all acoustic models consists of 26 English alphabet letters, augmented with the apostrophe and a word boundary token. To speed up training, we also use mixed-precision training.  We use the same SpecAugment and learning rate schemes as discussed above for STC models. We decay learning rate by a
factor of 2 each time the WER reaches a plateau on the validation sets. 

\subsection{Handwriting Recognition}

We use Adam Optimizer~\cite{kingma2014adam} with an initial learning rate of $0.02$ and run training using a  batchsize of 8 per GPU for all the experiments. All image are scaled to a maximum width, height of 600, 32 pixels respectively before feeding to the neural network. We use exponential learning rate decay schedule with a gamma factor of $10^{-1/90000} \approx 0.99997$. We use random projection transformations as the augmentation method while training. 

\section{A simple letter to word encoder}
\label{apx:l2w}
Since, we are working on direct-to-word models for ASR, a major hurdle would be to transcribe words not found in the training vocabulary. Recently, \cite{collobert2020word} propose an end-to-end model which outputs words directly, yet is still able to dynamically modify the lexicon with words not seen at training time. They use a convnet based letter to word embedding encoder which is jointly trained with the acoustic model and is able to accommodate new words at testing time.

In this work, we propose a simpler system for letter to word encoder which can be embedded into the acoustic model and thus avoiding the needs for a separate network. Consider $\gA_L$,  $\gA_W$ are the alphabets for letters, words respectively and $l_{max}$ is the maximum length of a word. We assume $\gA_L$ also contains two special letters - $c_{\text{\textit{blank}}}$ which produces \textit{blank} token required for CTC/STC models and  $c_{\text{\textit{pad}}}$  which is used to pad all words, \textit{blank} token to the same maximum word length $l_{max}$. We let the acoustic model produce a $(\lvert \gA_L \rvert
\times l_{max})$ dimensional vector for each time frame. 

\begin{figure}[ht]
\begin{tiny}
\begin{tikzpicture}
\matrix (B) [bmatrix] {
1 & 0 & 0 & 0 & 0 & \; 0 & 0 &0 & 0 & 1    & \; 0 & 0 & 0 &0 & 1 \\
0 & 0 & 1 & 0 &0 & \; 1 & 0 &0 & 0 & 0    & \; 0 & 1 & 0 &0 & 0 \\
0 & 0 & 1 & 0 &0 & \; 1 & 0 &0 & 0 & 0    & \; 0 & 0 & 0 &0 & 1 \\
0 & 0 & 0 & 1 & 0 &\; 0 & 0 &0 & 0 & 1    & \; 0 & 0 & 0 &0 & 1\\};
\node (eq) [right=1mm of B] {$\times$}; 
  
\node (eq1) [below of=B] {$\mE$};
\matrix (C) [bmatrix, right=1mm of eq] {
\;\;0.9 \\ -0.4 \\ \;\;0.2 \\  \;\;1.8 \\  -0.1 \\ \; \\  -0.4 \\  \;\;1.2  \\ -1.3 \\  \;\;0.1 \\  \;\;1.2 \\ \; \\  \;\;2.1  \\ -0.8 \\  -1.4 \\ \;\;0 \\ \;\;0.1 \\};
\node (eq1) [right=1mm of C,  text=Violet,text width = 0.1\textwidth] {a \\ b \\ c \\  $c_{\text{\textit{blank}}}$ \\  $c_{\text{\textit{pad}}}$ \\ [1ex] a\\  b \\ c \\ $c_{\text{\textit{blank}}}$  \\  $c_{\text{\textit{pad}}}$ \\ [1ex] a \\ b \\ c \\  $c_{\text{\textit{blank}}}$  \\  $c_{\text{\textit{pad}}}$ \\};
\node (eq) [right=1mm of eq1] {$=$};
\matrix (C) [bmatrix, right=1mm of eq] {\;\;2.3 \\
        -1.0 \\
        -0.3 \\
        \;\;3.1\\};
\node (eq) [right=1mm of C,   text=Violet,text width = 0.25\textwidth] {{a \\ cab \\ ca \\ [0.4ex]\textit{blank} }};
\end{tikzpicture}
\end{tiny}
\vskip -0.1in
\caption{Letter to word encoder in action. Matrix $\mE$ converts the letter scores over each timeframe to a score over words and \textit{blank}. $\gA_L = \{a, b, c,c_{\text{\textit{blank}}}, c_{\text{\textit{pad}}}\}$, $l_{max}=3$, $\gA_W = \{\text{a}, \text{cab}, \text{ca}\}$.  }
\label{fig:matmul}
\end{figure}

Since CTC/STC expect scores for $\gA_W$ and \textit{blank} for each time frame, we carefully construct a matrix $\mE$ of $1$s and $0$s which converts a vector of size  $(\lvert \gA_L \rvert \times l_{max} )$ to  $(\lvert \gA_W \rvert + 1) $ as shown in Figure~\ref{fig:matmul} . Since we are using $c_{\text{\textit{pad}}}$ token for padding, we can always assume every words and \textit{blank} token is a sequence of $l_{max}$ letters. Each row in $\mE$ is constructed by concatenating the one-hot representation of each letter  (including $c_{\text{\textit{pad}}}$) and it produces the score for a word or \textit{blank}. We use $\gA_L = \{a-z,\text{\textquotesingle},c_{\text{\textit{blank}}}, c_{\text{\textit{pad}}}\}$ and $l_{max}=20$ for all our experiments on LibriSpeech using words as the output tokens.

\begin{table*}[ht!]
\caption{WER comparison on LibriSpeech dev and test sets}
\label{ls-wer2}
\begin{center}
\begin{scriptsize}
\begin{sc}
\begin{tabular}{ccccccc}
\toprule
 \multirow{2}{*}{Method} & \multirow{2}{*}{Criterion} & Model Stride/ & Output &  \multirow{2}{*}{LM}  &  {Dev WER} & {Test WER}  \\
 & & Parameters & tokens & &  clean/other & clean/other \\
\midrule
 \multirow{2}{*}{Transformer} & \multirow{2}{*}{CTC} & \multirow{2}{*}{8/$\sim$ 300M} & \multirow{2}{*}{Words} &  - & 2.9/7.5 & 3.2/7.5 \\ 
 &&&  & Words 4-gram & 2.6/6.6 & 2.9/6.7 \\
 \tiny{\cite{collobert2020word}} & \multirow{2}{*}{seq2seq} &   \multirow{2}{*}{8/$\sim$ 300M} & \multirow{2}{*}{Words} &   - & 2.7/6.5 & 2.9/6.7 \\
  & & & & word 4-gram & 2.5/6.0 & 3.0/6.3 \\
\midrule
Transformer & \multirow{2}{*}{CTC} & \multirow{2}{*}{8/70M} & \multirow{2}{*}{Words} &  - & 2.3/6.8 & 2.8/7.0 \\ 
 \tiny{(Uses our simple letter to word encoder)} &&&  & Word 5-gram & 2.2/6.3 & 2.7/6.5\\ 
\bottomrule 
\end{tabular}
\end{sc}
\end{scriptsize}
\end{center}
\end{table*}

From Table~\ref{ls-wer2}, we can see that we are able to match the performance of the word based models from \citet{collobert2020word} using a much smaller acoustic model. Also, our system is simpler as we do not have a separate network for letter-to-word encoder. 

\section{Hyperparameters for token insertion penalty, $\lambda$}
\label{apx:tip}

In Table~\ref{tbl:lambda}, we report the hyperparameters used for training STC models on LibriSpeech (Table~\ref{ls-wer1}) and IAM (Table~\ref{iam-cer}).

\begin{table}[ht!]
\caption{Best performing hyperparameters used for token insertion penalty, $\lambda$ while training STC models for LibriSpeech (left) and IAM (right).}
\label{tbl:lambda}
\vskip 0.15in
\begin{center}
\begin{small}
\begin{sc}
\begin{minipage}[t]{.6\linewidth}
\centering
\begin{tabular}{cccc}
\toprule
Weak Label & \multirow{2}{*}{$p_0$} & \multirow{2}{*}{$p_{max}$} & \multirow{2}{*}{$t_{1/2}$} \\
Gen. Strategy \\ 
\midrule
\rule{0pt}{3ex} pDrop=0.1    & 0.1 & 0.3 & 8000 \\ 
\rule{0pt}{3ex} pDrop=0.4  & 0.4 & 0.7 & 8000 \\ 
\rule{0pt}{3ex} pDrop=0.7    & 0.5 & 0.9 & 8000 \\ 
\rule{0pt}{3ex} Split all samples into    & \multirow{4}{*}{0.3} & \multirow{4}{*}{0.6} &   \multirow{4}{*}{8000}       \\
3 parts randomly; \\ 
Assign pDrop=0.1,0.4,0.7 \\
for the splits \\ 
\rule{0pt}{3ex} Split all samples into    & \multirow{4}{*}{0.5} & \multirow{4}{*}{0.7} &   \multirow{4}{*}{8000} \\
3 parts randomly; \\ 
Assign pDrop=0.1,0.4,0.7 \\
for the splits \\ 
\bottomrule
\end{tabular}
\end{minipage} 
\begin{minipage}[t]{.3\linewidth}
\centering
\begin{tabular}{cccc}
\toprule
pDrop & $p_0$ & $p_{max}$ & $t_{1/2}$ \\
\midrule
0.1    & 0.5 & 0.8 & 10000 \\
0.3  & 0.5 & 0.9 & 10000 \\
0.5    & 0.5 & 0.9 & 10000 \\
0.7   & 0.7 & 0.9 & 10000 \\
\bottomrule

\end{tabular}

\end{minipage} 
\end{sc}
\end{small}
\end{center}
\vskip -0.1in
\end{table}


\end{document}